\documentclass[runningheads]{llncs}
\usepackage[T1]{fontenc}
\usepackage{graphicx}

\usepackage{amsmath}
\usepackage{cleveref}
\usepackage{amsfonts}
\usepackage{color}
\usepackage{xcolor}
\usepackage{amssymb}
\usepackage{url}
\usepackage{float}
\usepackage{siunitx}

\DeclareMathOperator*{\argmin}{arg\,min}

\usepackage{dsfont}
\newcommand{\indicator}{\mathds{1}}
\newcommand{\smallFixM}{\emph{uni\_sm}}
\newcommand{\fixN}{\emph{uni\_fix\_n}}

\begin{document}
\title{Distributed Multi-Coverage for Robot Swarms}
%
%
\author{Mariem Guitouni\orcidID{0009-0000-0020-6314} \and
Aaron T. Becker\orcidID{0000-0001-7614-6282}}
\authorrunning{Guitouni \and Becker}
%
\institute{University of Houston, Texas, USA, 77204
\email{\{mguitoun,atbecker\}@cougarnet.uh.edu}}
\maketitle              
\begin{abstract}
Autonomous drone swarms deployed for surveillance, environmental monitoring, and infrastructure inspection must maintain reliable coverage of critical assets despite robot failures. This requires multi-coverage: each asset must be observed by multiple robots for redundancy, with coverage requirements varying by asset importance. While recent work~\cite{guitouni2025multi} has solved the centralized problem optimally using integer programming, practical deployments face constraints that demand distributed solutions: robots operate with limited communication ranges, onboard computation restricts global planning, and partial system failures must not cause mission abort. We present a distributed multi-coverage algorithm for robot swarms operating with local sensing, local communication, and no global coordination\footnote{Code 
available at: \url{https://doi.org/10.5281/zenodo.18626854}}.

\end{abstract}
\section{Introduction}\label{sec:int}
\begin{figure}[t]
\centering
\includegraphics[width=\linewidth]{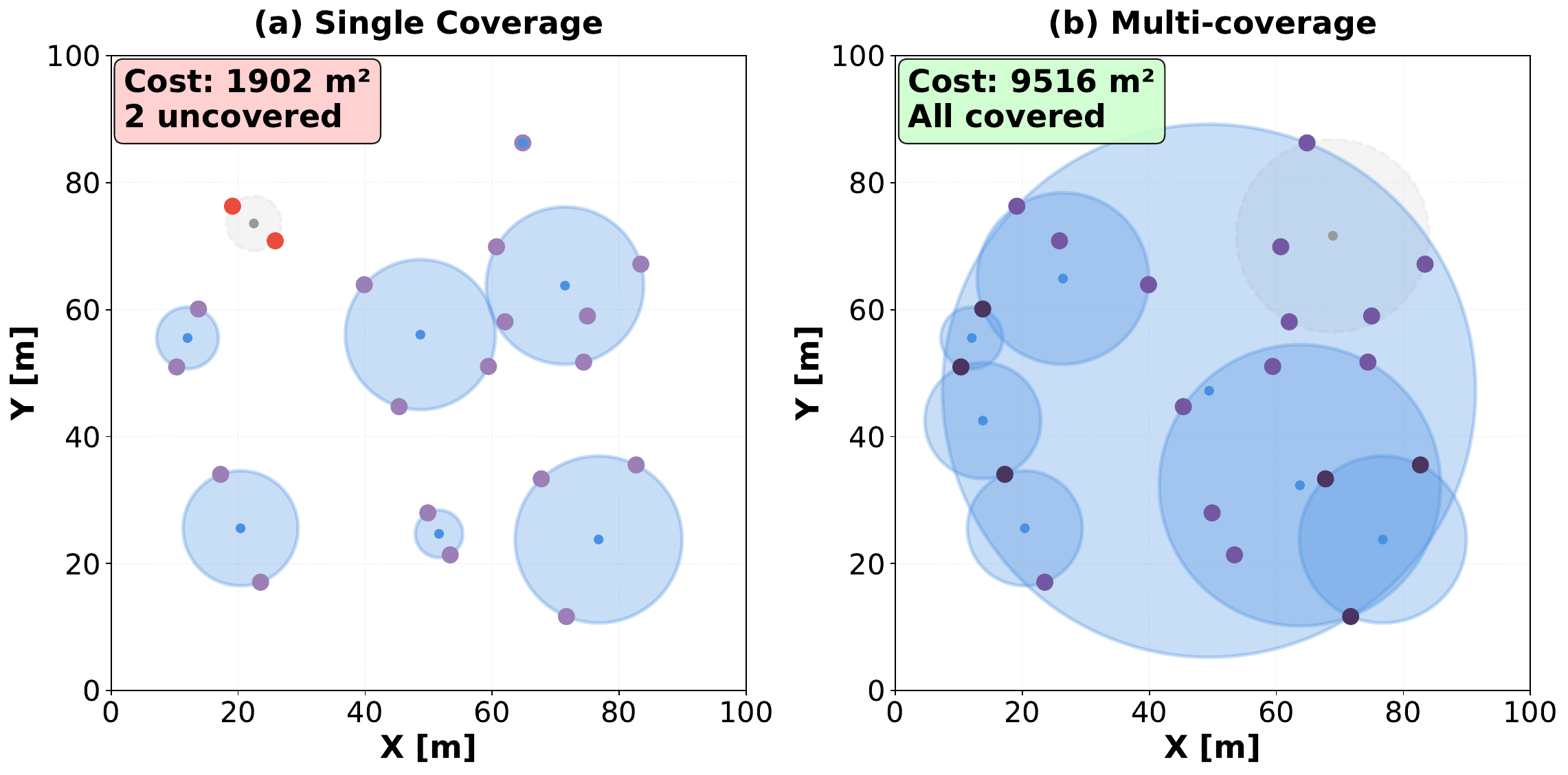}
\caption{\small Multi-coverage enables fault tolerance: (a) single coverage ($\kappa=1$) fails when a robot is lost, while (b) heterogeneous multi-coverage maintains monitoring of critical assets. Asset coverage requirements: \textcolor[HTML]{9B7FB6}{$\bullet$} $\kappa = 1$, \textcolor[HTML]{7457A3}{$\bullet$} $\kappa = 2$, \textcolor[HTML]{4A3560}{$\bullet$} $\kappa = 3$, \textcolor[HTML]{E74C3C}{$\bullet$} indicates uncovered assets. Robot coverage disks: \textcolor[HTML]{4A90E2}{\rule{0.8em}{0.8em}} active robots, \textcolor[HTML]{CCCCCC}{\rule{0.8em}{0.8em}} failed robots. ($n=20$ assets, $m=8$ robots in a \qty{100}{\meter} $\times$ \qty{100}{\meter} workspace.)}
\label{fig:motivation}
\end{figure}

The optimal placement of robots to monitor a set of assets is a fundamental problem in robotics, with applications ranging from environmental monitoring to infrastructure inspection. When robots are subject to failures, single-coverage is insufficient---each asset must be observed by multiple robots. Figure~\ref{fig:motivation} illustrates how multi-coverage provides fault tolerance when robots fail. This leads to the \emph{General Multi-Coverage }problem: given $n$ assets in a workspace and $m$ robots with circular sensing regions, determine positions $\mathbf{y}_i \in \mathbb{R}^2$ and radii $r_i \geq 0$ such that each asset $p$ is covered by at least $\kappa(p)$ robots, while minimizing the total sensing cost $\pi$$\sum_{i=1}^m r_i^2$, which relates directly to energy consumption.

The centralized variant of this problem has recently been solved to provable optimality using Integer Programming~\cite{guitouni2025multi}. By exploiting geometric properties---that any optimal disk covering a subset of points must have 1, 2, or 3 points on its boundary---a finite candidate set of $O(n^3)$ disk positions can be enumerated. Coverage constraints are then formulated as integer linear constraints, yielding provably optimal solutions via modern IP solvers. However, this approach inherently assumes global knowledge: a central planner must know all asset locations, compute all candidate positions, and communicate the solution to all robots.

This assumption fails in many realistic deployment scenarios. Consider a swarm of $m = 200$ drones monitoring a wildfire with $n = 500$ hotspots~\cite{chen2022wildland}. The centralized IP formulation generates $O(500^3) \approx 125$ million candidate disks, far exceeding computational tractability. Alternatively, consider a drone swarm inspecting critical structural points---welds, joints, and stress points on a large bridge structure spanning \qty{1.2}{\kilo\meter}, ~\cite{ASCE_AnIconAt80_2017}. With typical WiFi-based communication ranges of approximately \qty{100}{\meter}~\cite{fanet_hybrid_comm}, further degraded by metal structural interference from bridge components, and with drones continuously repositioning for inspection, maintaining a global communication topology for centralized coordination becomes impractical.

These motivating scenarios share common constraints: \emph{local sensing} (each robot observes only assets within radius $r_{\max}$), \emph{local communication} (robots exchange information only with neighbors within range $r_{\text{comm}}$), and \emph{distributed computation} (no central processor coordinates the swarm). Under these constraints, centralized algorithms are inapplicable.

We present a distributed algorithm for multi-coverage with an \emph{exploration phase} where robots utilize Lloyd's algorithm initialized from a structured grid to discover all assets through grid-based partitioning, followed by an \emph{optimization phase} where robots ensure coverage requirements $\kappa(p)$ are satisfied using only local information about nearby assets and neighboring robot states and they minimize their coverage costs during \emph{refinement phase}.

Our technical contributions are:
\begin{enumerate}
    \item A distributed multi-coverage algorithm, operating under local sensing and communication constraints without global coordination.
    \item Experimental characterization across static and dynamic scenarios, quantifying fundamental tradeoffs between solution optimality, computational scalability, and adaptability to changing environments.
\end{enumerate}

The remainder of this paper is organized as follows. Section~\ref{sec:literature} reviews related work on multi-coverage optimization, and distributed control. Section~\ref{sec:formulation} provides formal problem definition. Section~\ref{sec:method} presents our distributed algorithm. Section~\ref{sec:results} evaluates performance through comprehensive experiments comparing centralized and distributed approaches. Section~\ref{sec:discussion} discusses the obtained results and concludes.

\section{Related Work}\label{sec:literature}
The distributed multi-coverage problem intersects two major research streams: optimization methods for multi-coverage with quality guarantees, and distributed control algorithms enabling decentralized coordination. We review recent advances in both areas and position our contribution.

\subsection{Multi-Coverage Optimization}
The multi-coverage problem---ensuring each asset is monitored by multiple robots for fault tolerance---has been extensively studied under centralized coordination. Early work by Slijepcevic and Potkonjak~\cite{slijepcevic2001power} introduced power-efficient organization through set covers that achieve $k$-coverage, where each field is monitored by $k$ robots. Huang and Tseng~\cite{huang2005coverage} formalized the coverage problem as a decision problem and presented polynomial-time algorithms to verify whether every region is covered by at least $k$ robots, establishing theoretical foundations for coverage verification. Their work demonstrated applications in fault-tolerant monitoring and energy conservation through redundant sensor scheduling.
Our prior work~\cite{guitouni2025multi} solved the centralized variant optimally using Integer Programming. However, this centralized approach inherently requires global knowledge of all asset positions and computational complexity that grows cubically with problem size, limiting scalability for large deployments.
Recent metaheuristic approaches provide practical alternatives when exact methods become intractable. Zhou et al.~\cite{zhou2024sensor} applied hybrid Genetic Algorithm and Particle Swarm Optimization to visual sensor placement, balancing reconstruction error and coverage quality. More recently, a PSO-based deployment framework~\cite{pso_deployment2025} achieved 91.4\% average coverage with extended operational lifetime for heterogeneous wireless sensor networks. While these methods demonstrate strong empirical performance, they remain centralized and cannot adapt locally to dynamic environmental changes or sensor failures during mission execution.

\subsection{Distributed Coverage}
Distributed control enables scalability and robustness by allowing agents to make decisions using only local information.
Centroidal Voronoi Tessellation (CVT) provides the theoretical foundation for distributed coverage. The seminal work of Cort{\'e}s et al.~\cite{cortes2004coverage} introduced distributed control laws that partition the workspace among agents using Voronoi diagrams, with each agent moving toward its cell's centroid weighted by an importance function. Using Lyapunov analysis, they proved convergence to locally optimal configurations that minimize a coverage metric. Recent extensions have addressed CVT limitations. Lee et al.~\cite{lee2015multirobot} introduced time-varying density functions enabling dynamic adaptation to changing environments. Zhou et al.~\cite{zhou2017fast} proposed Buffered Voronoi Cells (BVC) that maintain safety margins between agents, enabling collision-free coordination for dynamic vehicles. While CVT methods excel at continuous positioning optimization, they optimize where agents position themselves to minimize a coverage metric over continuous space, whereas multi-coverage requires ensuring discrete coverage requirements $\kappa(p)$ are satisfied through geometric disk placement with optimized radii. 

\section{Problem Formulation}\label{sec:formulation}
We formalize the distributed multi-coverage problem, extending the centralized formulation from~\cite{guitouni2025multi} with realistic constraints on sensing, communication, and coordination. 

\subsection{ Multi-Coverage Problem}

Given a set $S$ of $n$ assets (points) in a bounded workspace $\mathcal{W} \subset \mathbb{R}^2$, and $m$ robots indexed by $i \in \{1, \ldots, m\}$, the \emph{General Multi-Coverage (GMC)} problem seeks to assign each robot $i$ a position $\mathbf{y}_i \in \mathcal{W}$ and sensing radius $r_i \geq 0$ such that each asset $p \in S$ is covered by at least $\kappa(p)$ robots, where $\kappa: S \to \mathbb{N}$ specifies the coverage requirement for each asset based on its importance, while minimizing the total sensing cost $\pi\sum_{i=1}^m  r_i^2$, which directly relates to energy consumption.
Formally, the problem is defined as follows

\begin{align}
    \text{minimize} \quad & \pi\sum_{i=1}^m  r_i^2 \label{eq:objective}\\
    \text{subject to} \quad & \sum_{i=1}^m \indicator[\|\mathbf{y}_i - p\| \leq r_i] \geq \kappa(p), \quad \forall p \in S \label{eq:coverage}\\
    & r_i \geq 0, \quad \mathbf{y}_i \in \mathcal{W}, \quad \forall i \in \{1, \ldots, m\} \,.\label{eq:domain}
\end{align}
Here $\indicator[\cdot]$ is the indicator function. The centralized solution~\cite{guitouni2025multi} assumes global knowledge: a central planner knows all asset positions, computes all $O(n^3)$ candidate disk positions, and optimally assigns robots using integer programming.

\subsection{Distributed Constraints}

In practical deployments, robots operate under strict locality constraints that preclude centralized coordination. We introduce three fundamental restrictions:

\paragraph{Local Sensing.} Each robot $i$ can only detect assets within a maximum sensing radius $r_{\max} > 0$. While a robot may choose any radius $r_i \in [0, r_{\max}]$, it cannot observe assets beyond this physical limit:
\begin{equation}
    r_i \leq r_{\max}, \quad \forall i \in \{1, \ldots, m\} \label{eq:max_radius} \,.
\end{equation}

\paragraph{Local Communication.} Robots exchange information only with neighbors within communication range $r_{\text{comm}} > 0$. For robot $i$ at position $\mathbf{y}_i$, the set of neighbors is:
\begin{equation}
    \mathcal{N}_i = \{j \in \{1, \ldots, m\} : j \neq i \text{ and } \|\mathbf{y}_i - \mathbf{y}_j\| \leq r_{\text{comm}}\}   \,. \label{eq:neighbors}
\end{equation}
Robot $i$ can only access the state $(\mathbf{y}_j, r_j)$ and covered asset list of robots $j \in \mathcal{N}_i$.

\paragraph{No Global Coordination.} Robots have no knowledge of the global state. Each robot makes decisions based solely on:
\begin{itemize}
    \item Its own state $(\mathbf{y}_i, r_i)$ and locally sensed assets $A_i \subseteq S$.
    \item Neighbor $\mathcal{N}_i$ within communication range and their states $\{(\mathbf{y}_j, r_j, A_j) : j \in \mathcal{N}_i\}$.
    \item Global parameters: workspace bounds $\mathcal{W} = [x_{\min}, x_{\max}] \times [y_{\min}, y_{\max}]$, coverage requirements $\kappa(\cdot)$, and number of robots $m$.
\end{itemize}

\section{Method}\label{sec:method}

We present a distributed algorithm for multi-coverage that operates in three phases: (1) \emph{exploration} to discover assets through systematic deployment, (2) \emph{optimization} to satisfy coverage requirements through local coordination, and (3) \emph{refinement} to minimize sensing cost by reducing overcoverage. Throughout all phases, when a robot recomputes its minimum bounding disk using Welzl's algorithm~\cite{welzl1991smallest}, it updates both position $\mathbf{y}_i$ (disk center) and radius $r_i$ (disk radius), physically moving to the new center location.

\subsubsection{Phase 1: Exploration}
Robots utilize Lloyd's algorithm~\cite{lloyd1982least} to distribute themselves optimally across the workspace. Initial centroids are determined by a structured grid partition that minimizes wasted cells:
\begin{equation}
    (n_r, n_c) = \argmin_{\substack{n_r, n_c \in \mathbb{N} \\ n_r \cdot n_c \geq m}} \left(n_r \cdot n_c - m\right) + \lambda |n_r - n_c|  \,,
\end{equation}
where $\lambda > 0$ favors square grids. Robot $i$ initializes at cell $(r, c) = (\lfloor i / n_c \rfloor, i \bmod n_c)$:
\begin{equation}
    \mathbf{y}_i^{(0)} = \begin{pmatrix} x_{\min} + (c + 0.5)\mathrm{\Delta} x \\ y_{\min} + (r + 0.5)\mathrm{\Delta} y \end{pmatrix}  \,,
\end{equation}
where $\mathrm{\Delta} x = (x_{\max} - x_{\min})/n_c$ and $\mathrm{\Delta} y = (y_{\max} - y_{\min})/n_r$.

Lloyd's algorithm iteratively refines positions via Voronoi partitioning. 
During each iteration, robots sense nearby assets within range $r_{\text{max}}$ 
and construct Voronoi cells by assigning each sensed asset to the nearest 
robot position. Each robot then relocates to the centroid of its assigned 
assets:
\begin{equation}
    \mathbf{y}_i^{(t+1)} = \frac{1}{|V_i^{(t)}|} \sum_{j \in V_i^{(t)}} \mathbf{p}_j  \,,
\end{equation}
where $V_i^{(t)} = \{j : \|\mathbf{p}_j - \mathbf{y}_i^{(t)}\| \leq \|\mathbf{p}_j - \mathbf{y}_k^{(t)}\|, \forall k \neq i\}$ is robot $i$'s Voronoi cell. The radius is set to $r_i^{(t+1)} = \min(\max_{j \in V_i^{(t)}} \|\mathbf{p}_j - \mathbf{y}_i^{(t+1)}\|, r_{\max})$. Iteration terminates when $\max_i \|\mathbf{y}_i^{(t+1)} - \mathbf{y}_i^{(t)}\| < tol$ , with $tol= 0.01.$

After Lloyd convergence, each robot computes the minimum bounding disk of its covered assets using Welzl's algorithm~\cite{welzl1991smallest}.

\subsubsection{Phase 2: Optimization}
Robots coordinate through local communication to achieve required coverage levels. Each robot $i$ exchanges its covered assets list with neighbors $\mathcal{N}_i$ to compute local coverage counts:
\begin{equation}
    c_i(p) = \indicator[p \in A_i] + \sum_{j \in \mathcal{N}_i} \indicator[p \in A_j]  \, , \label{eq:local_coverage}
\end{equation}
where $\indicator[\cdot]$ is the indicator function.
For each undercovered asset $p$ where $c_i(p) < \kappa(p)$ and $p \notin A_i$, robot $i$ computes the marginal cost (area increase) of covering $p$:
\begin{equation}
    \mathrm{\Delta}_i(p) = \pi r_i^{\text{new}}(p)^2 - \pi r_i^2  \,,\label{eq:marginal_cost}
\end{equation}
where $r_i^{\text{new}}(p)$ is the radius of the minimum bounding disk covering $A_i \cup \{p\}$. If $r_i^{\text{new}}(p) > r_{\max}$, set $\mathrm{\Delta}_i(p) = +\infty$. After exchanging marginal costs with neighbors, robot $i$ selects the minimum-cost robot for each undercovered asset:
\begin{equation}
    j^* = \argmin_{j \in \mathcal{N}_i \cup \{i\}} \mathrm{\Delta}_j(p)  \, , \label{eq:winner}
\end{equation}
Ties within relative tolerance $\epsilon = 0.01$ are broken deterministically via hash function $h(\text{iter}, p, j)$ to ensure reproducibility. If $i = j^*$, robot $i$ adds $p$ to $A_i$ and recomputes its minimum bounding disk, repositioning itself to the new disk center and adjusting its radius 
accordingly. If no progress occurs (all costs infinite), the robot with maximum capacity $r_{\max} - r_i$ covers its nearest undercovered asset. This process iterates until all coverage requirements are satisfied. The algorithm operates in synchronous rounds where all robots simultaneously exchange information with neighbors, compute marginal costs, determine winners via \eqref{eq:winner}, and update their disk configurations before proceeding to the next iteration.

After achieving required coverage, robots perform pairwise asset swaps with neighbors to reduce total sensing cost~\cite{shahsavar2025tracking}. For each neighbor pair $(i,j) \in \mathcal{N}_i$ where $i < j$, robot $i$ considers transferring asset $p \in A_i$ to robot $j$ if: (1) $\|\mathbf{x}_p - \mathbf{y}_j\| < \|\mathbf{x}_p - \mathbf{y}_i\|$ (asset closer to receiving robot), (2) $\|\mathbf{x}_p - \mathbf{y}_i\| > 0.9 r_i$ (asset near donor's boundary), and (3) $c_i(p) - \indicator[p \in A_j] \geq \kappa(p)$ (coverage preserved after swap). Both forward and reverse swaps between each pair are evaluated. The swap is executed if it reduces total area by threshold $\tau = 0.005$: $\pi(r_i^2 + r_j^2) - \pi(r_i'^2 + r_j'^2) > \tau \cdot \pi(r_i^2 + r_j^2)$, where $r_i'$ and $r_j'$ are the radii of minimum bounding disks covering $A_i \setminus \{p\}$ and $A_j \cup \{p\}$ respectively, subject to $r_i', r_j' \leq r_{\max}$. The process iterates until no beneficial swaps exist.

\subsubsection{Phase 3: Refinement}
To minimize sensing cost, each robot $i$ iteratively removes overcovered assets from $A_i$ where $c_i(p) - 1 \geq \kappa(p)$ (using local coverage $c_i(p)$ from~\eqref{eq:local_coverage}) and recomputes a smaller bounding disk. Before removing asset $p$, robot $i$ verifies $c_i(p) - 1 - |\{j \in \mathcal{N}_i : j \text{ removing } p\}| \geq \kappa(p)$ to prevent simultaneous removals by neighbors from violating coverage constraints. 
This continues until no further radius reduction is possible. Because robots use only local neighbor information, the result may retain some overcoverage compared to the globally-optimal solution. A robot cannot determine if an asset is 
overcovered beyond its communication neighborhood—if additional robots outside 
$r_{\text{comm}}$ also cover the asset, this information is unavailable for 
local decision-making.

\section{Results}\label{sec:results}
\subsubsection{Experimental Setup}
For performance evaluation, we use benchmark instances from~\cite{guitouni2025multi}.
These instances consist of two configurations: \textit{uni\_sm} with variable asset count ($n = 20$ to $200$) and fixed robot count ($m = 20$), and \textit{uni\_fix\_n} with fixed asset count ($n = 250$) and variable robot count ($m = 20$ to $100$). Assets are uniformly distributed in a $100\,\text{m} \times 100\,\text{m}$ workspace with coverage requirements $\kappa(p)$ uniformly sampled from $\{1, 2, 3\}$, representing heterogeneous importance levels. 
We set communication range $r_{\text{comm}}=$ \qty{55}{\meter} and maximum sensing radius $r_{\text{max}}=$ \qty{40}{\meter}. 
All experiments were run on an Intel Core Ultra 7 165H with 32 GB RAM.

\subsubsection{Convergence Behavior}
\begin{figure}[h]
    \centering
    
    \includegraphics[width=1\textwidth]{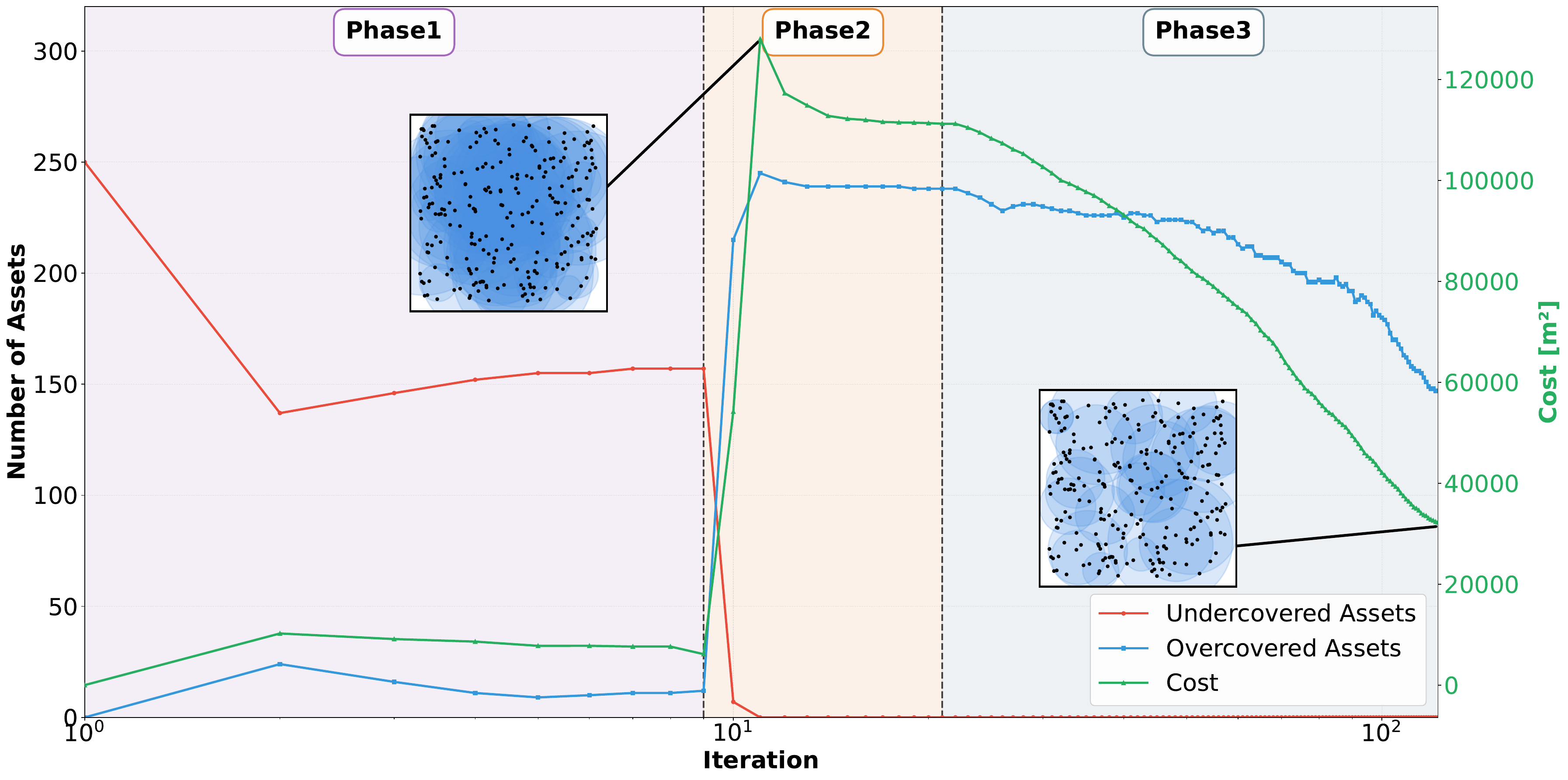}
    \caption{\small Convergence behavior of the distributed algorithm showing the number of undercovered/overcovered assets (left axis) and total coverage cost (right axis) across three phases. Inset visualizations show robot coverage disks (blue circles) centered at robot positions, with black dots representing asset locations that robots must cover. Phase 1 explores the workspace, Phase 2 ensures coverage requirements are met, and Phase 3 optimizes cost by reducing overcoverage. Parameters: $n=250$ assets, $m=50$ robots, $r_{\text{comm}}=$ \qty{55}{\meter}, $r_{\text{max}}=$ \qty{40}{\meter}, in a workspace \qty{100}{\meter} $\times$ \qty{100}{\meter}. For all lines, lower is better.}
    \label{fig:dist.convergence}
    
\end{figure}

Fig.~\ref{fig:dist.convergence} illustrates the convergence behavior of the distributed algorithm across its three phases. Phase 1 rapidly deploys robots using Lloyd's algorithm to explore the workspace establishing initial coverage. Phase 2 eliminates all undercovered assets through distributed coordination, achieving full coverage while overcoverage increases to 240 assets and cost rises to \qty{128079.89}{\meter\squared}. Phase 3 gradually reduces overcoverage to 147 assets while maintaining zero undercovered assets, achieving a cost reduction to \qty{32258.03}{\meter\squared}.

\subsubsection{ Performance Evaluation}
\begin{figure}[h]
   
    \includegraphics[width=\textwidth]{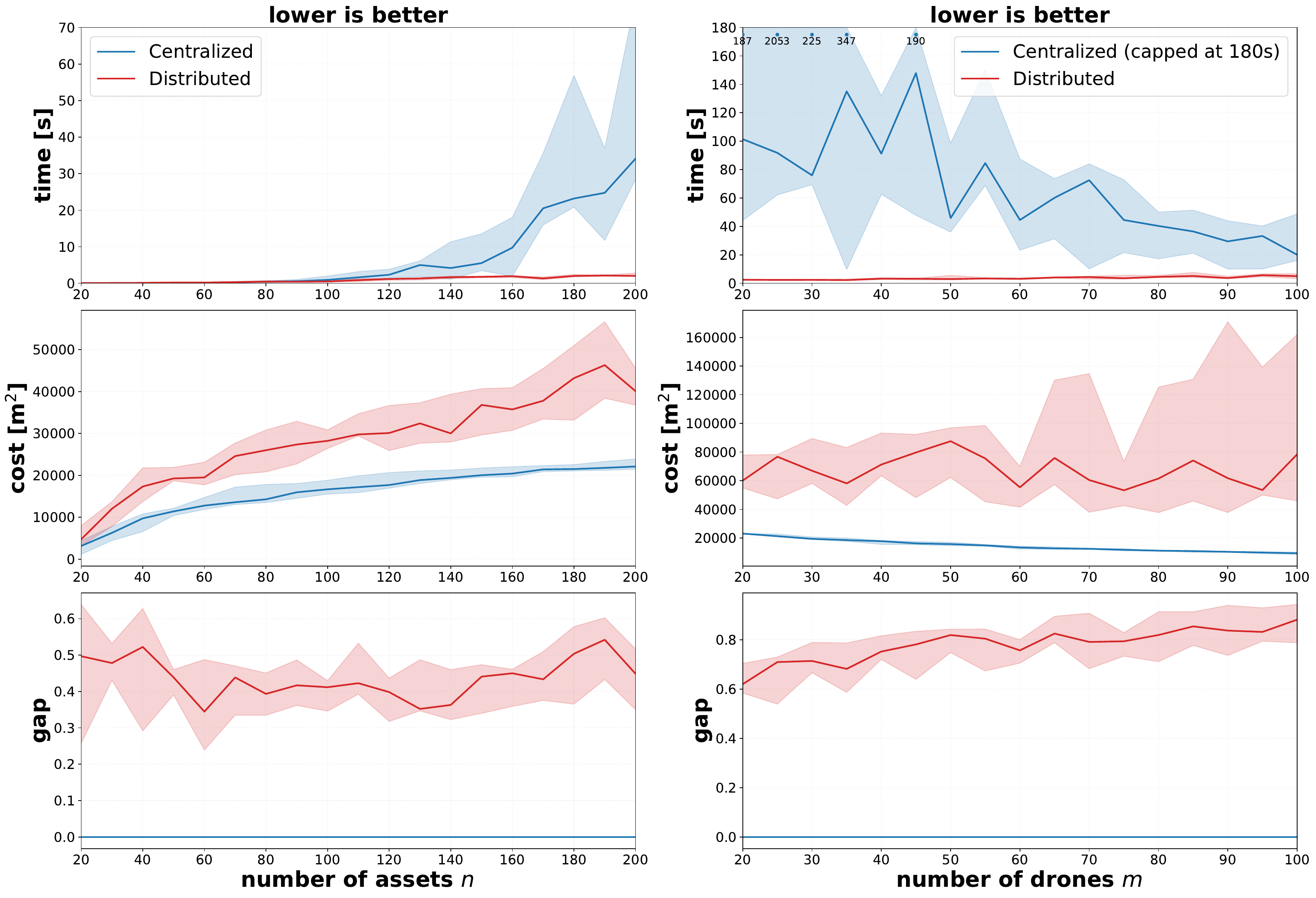}

    \caption{\small \label{fig:2d} 
     Comparison of runtime, coverage cost (total area), and optimality gap between the optimal centralized solver \cite{guitouni2025multi} and the decentralized algorithm. Lines show median over 5 trials; shaded regions show min-max range. On all plots, lower is better. (Left) \smallFixM{}: fixed $m=20$, variable $n$ (Right) \fixN{}: fixed $n=250$, variable $m$. Instance details are provided in the Experimental Setup.}
        
\end{figure}

Fig.~\ref{fig:2d} compares our distributed algorithm against the centralized IP baseline, which computes the global optimum. The distributed algorithm maintains near-constant computation time ($<$ \qty{5}{\second}) regardless of problem size, achieving median time-to-feasibility of 0.05 s and time-to-final-refinement of 0.61 s 
for uni\_sm , 0.19 s and 3.43 s for uni\_fix\_n. While centralized time grows up to \qty{60}{\second} for \smallFixM{} and reaches \qty{2053}{\second} for \fixN{}, yielding speedups of $12\times$ and $410\times$ respectively. As expected, the local optimization approach incurs an optimality gap compared to the global optimum: on average 45.78\% for \smallFixM{} and 69.41\% for \fixN{}. The gap increases with robot count due to increased coordination complexity in local decision-making. Nevertheless, all distributed solutions satisfy coverage requirements $\kappa(p)$. Applications requiring real-time response and operating under local sensing and communication constraints justify accepting higher cost rather than globally optimal solutions, while centralized optimization remains preferable for offline planning with relaxed time constraints and global control possibility.

\subsubsection{Centralized Failure Scenario}

\begin{figure}[htb]

    \centering
    \includegraphics[width=0.48\textwidth]{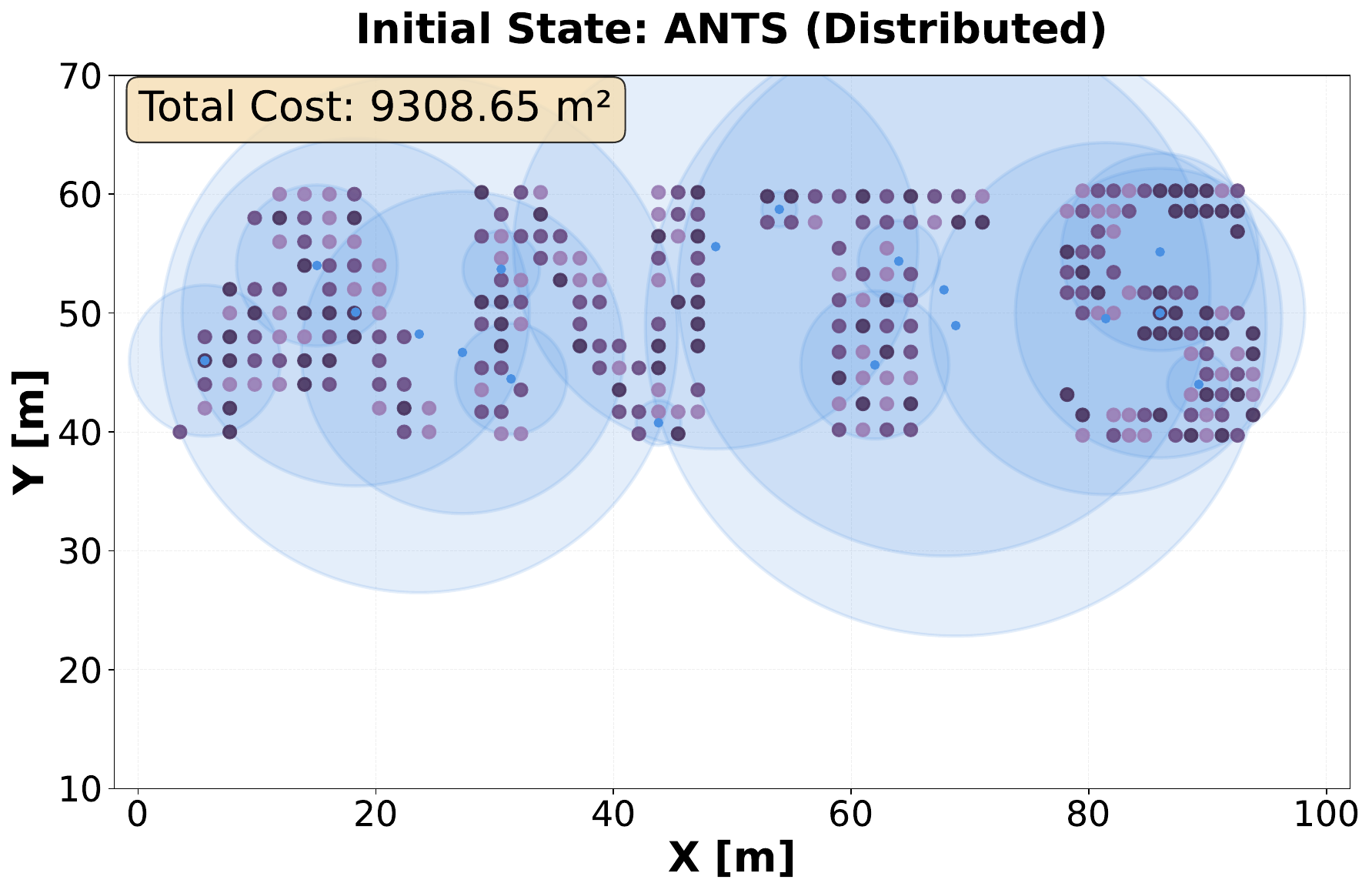}
    \includegraphics[width=0.48\textwidth]{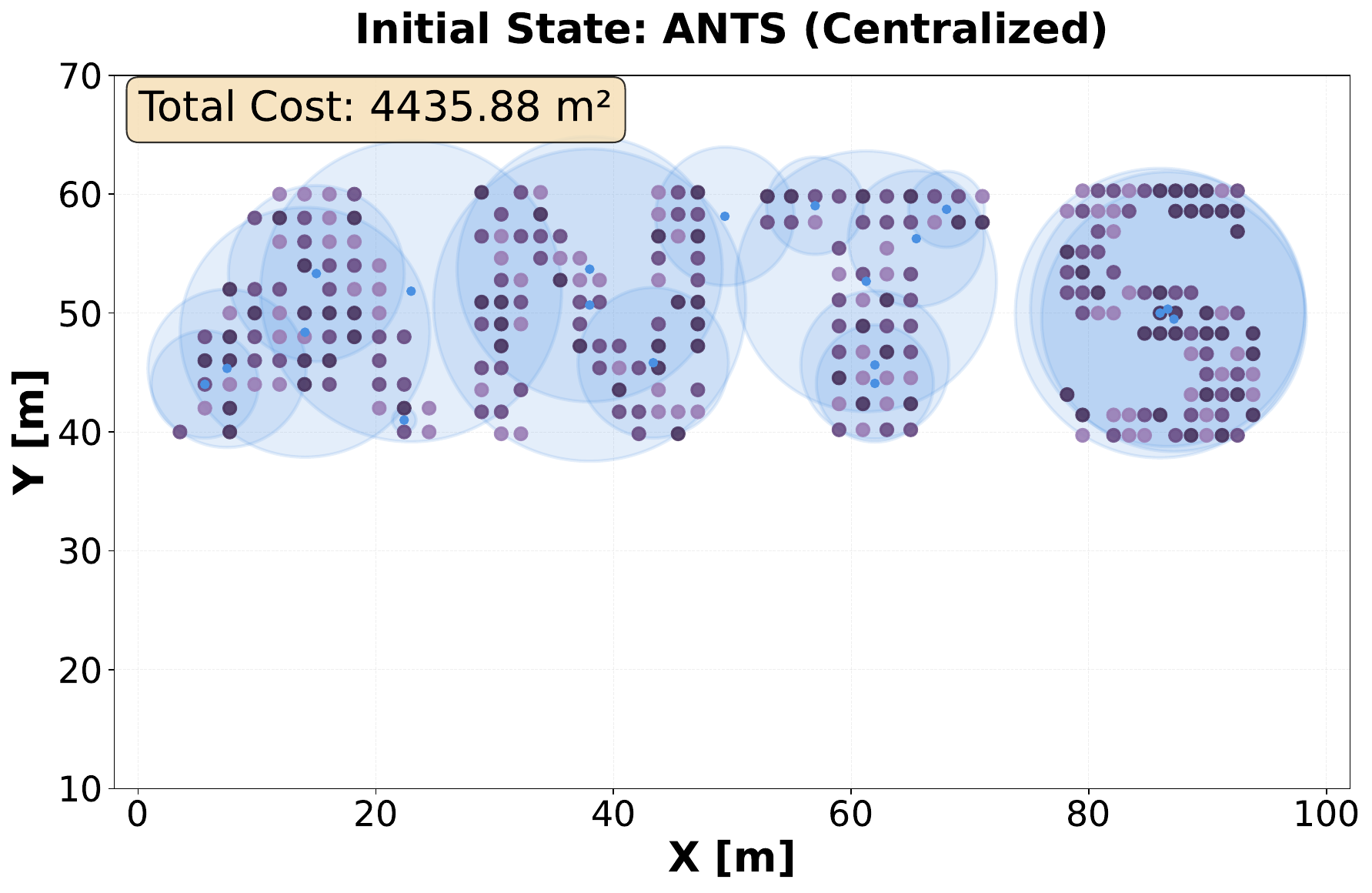}
        (a) Distributed: Initial coverage. ~~~~~~ (b) Centralized: Initial coverage.
    \includegraphics[width=0.48\textwidth]{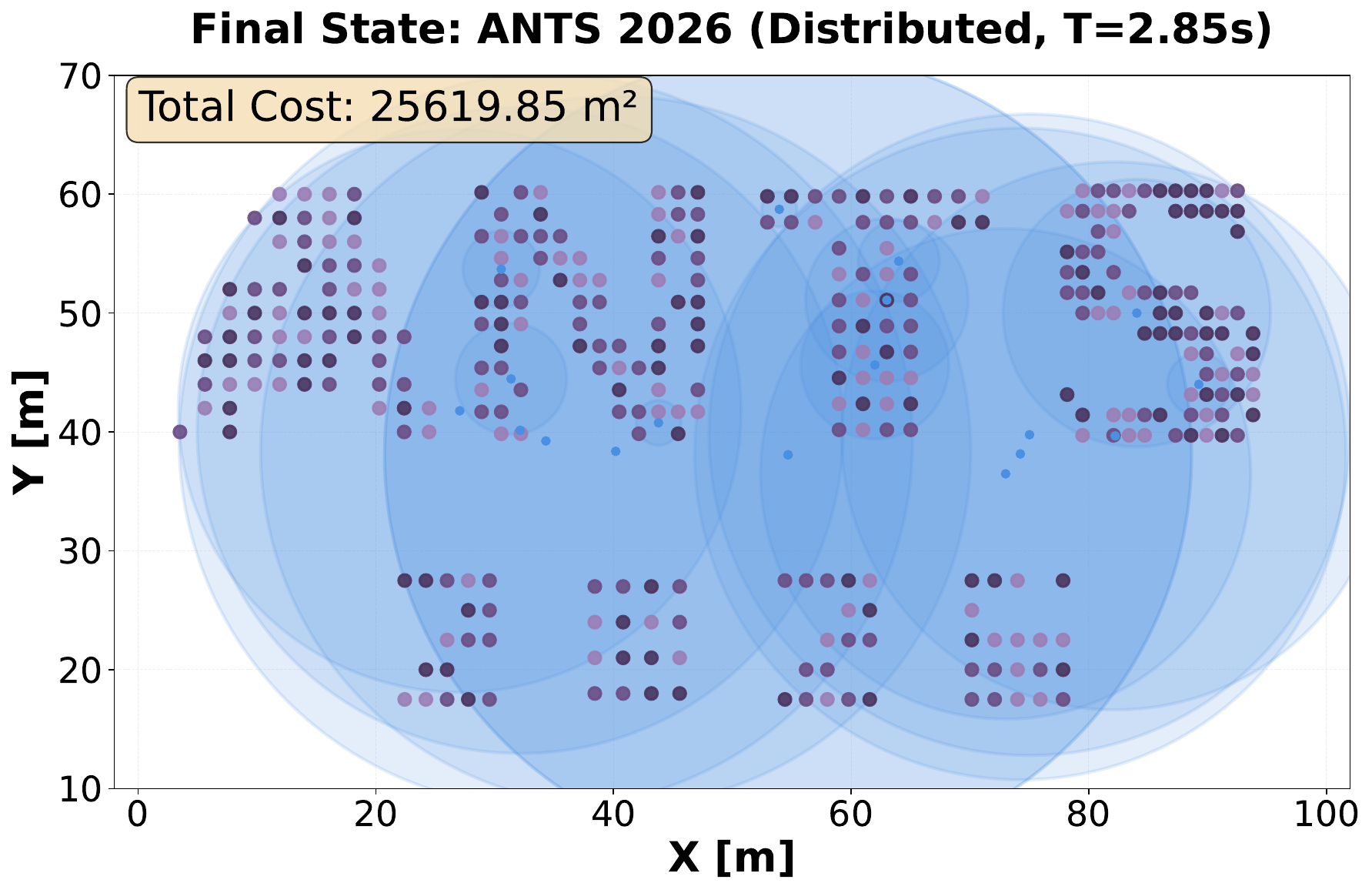}
    \includegraphics[width=0.48\textwidth]{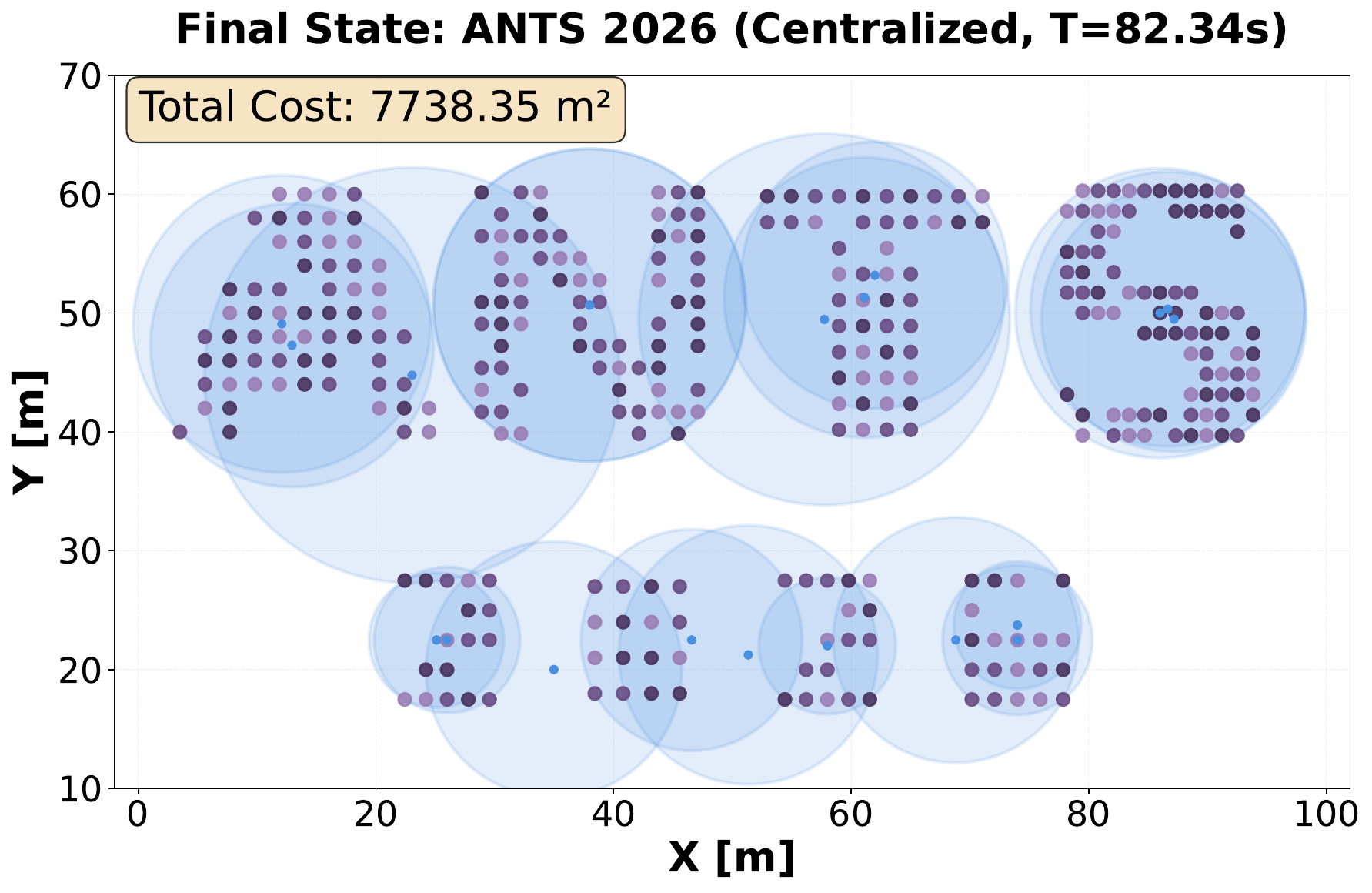}  
    
        ~~~(c) Distributed: Local adaptation.   ~~~~~~  (d) Centralized: Global recomputation.

        
        
    
    \caption{\small \label{fig:dynamic_comparison}Comparison of distributed vs.\ centralized approaches for dynamic asset appearance. The workspace contains 270 initial ANTS assets and 70 new 2026 assets (340 total), covered by 20 robots. Asset colors indicate coverage requirements: 
\textcolor[HTML]{9B7FB6}{$\bullet$} $\kappa = 1$, 
\textcolor[HTML]{6B4F86}{$\bullet$} $\kappa = 2$, 
\textcolor[HTML]{4A3560}{$\bullet$} $\kappa = 3$. Translucent blue disks show robot coverage regions.}

\end{figure}

To demonstrate operational advantages under dynamic conditions, we design an experiment where new assets appear during an active mission. We deploy $m=20$ robots to cover $n_{\text{init}}=270$ assets forming ``ANTS'' in a \qty{100}{\meter} $\times$ \qty{100}{\meter} workspace, with coverage requirements $\kappa(p) \in \{1,2,3\}$ and sensing parameters $r_{\max}=$ \qty{40}{\meter}, $r_{\text{comm}}=$ \qty{55}{\meter}. After both approaches achieve initial coverage, $n_{\text{new}}=70$ additional assets forming ``2026'' appear at $T=0$, simulating scenarios such as wildfire spread or structural damage discovery. The distributed algorithm responds through local adaptation: nearby robots detect new assets, trigger Phase~2 optimization using local communication, and adjust positions/radii via marginal cost bidding while distant robots remain unaffected. The centralized solver must recompute the entire solution, solving a new IP with $O(340^3) \approx 39.3$ million candidate disks and needs \qty{82.34}{\second}. Figure~\ref{fig:dynamic_comparison} shows both approaches across initial coverage and post-adaptation states for the combined $n_{\text{total}}=340$ asset system.

\subsubsection{Sensitivity Analysis}
\begin{figure}[h]
    \centering
         \includegraphics[width=0.48\textwidth]{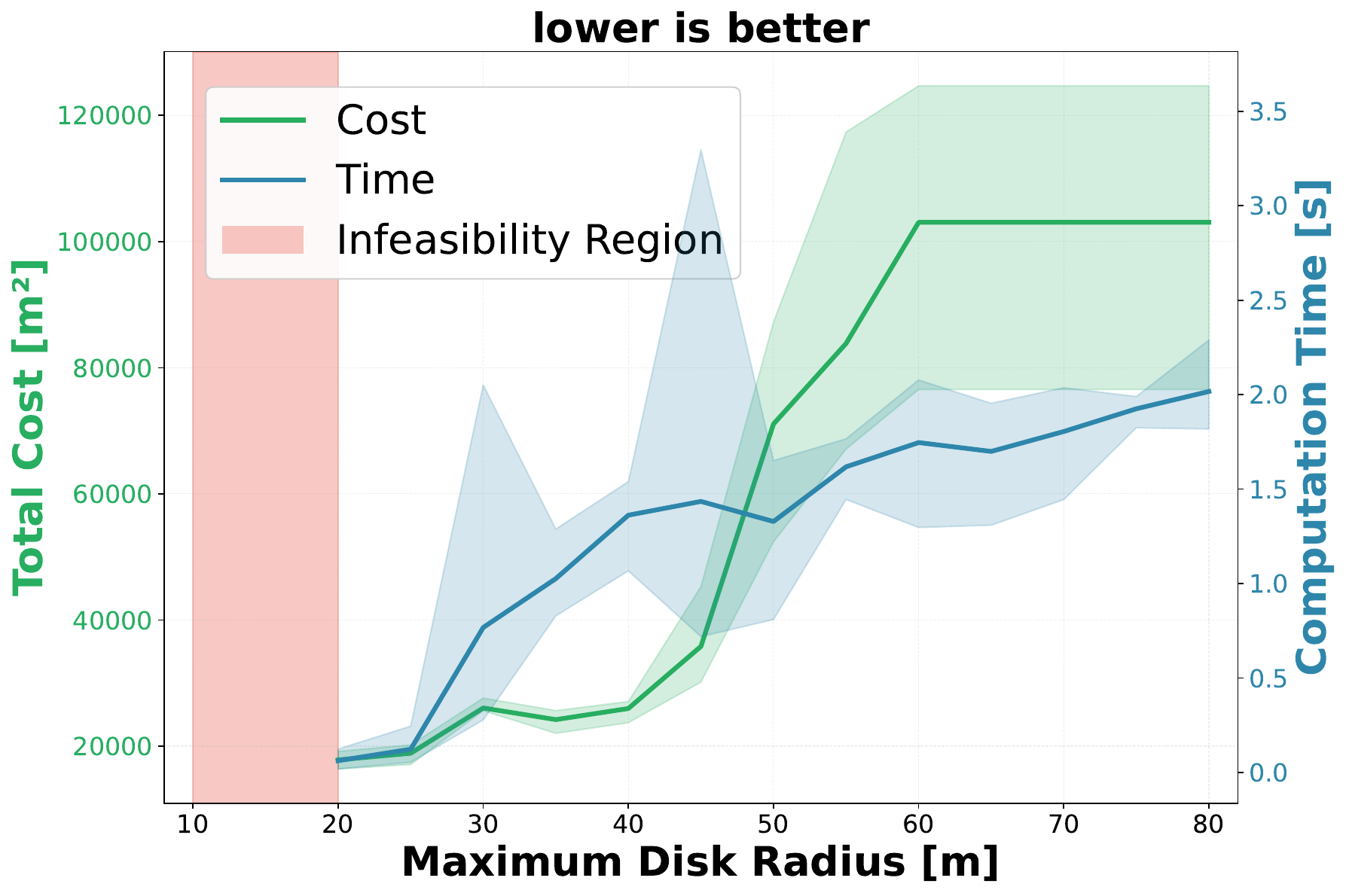}
         \includegraphics[width=0.48\textwidth]{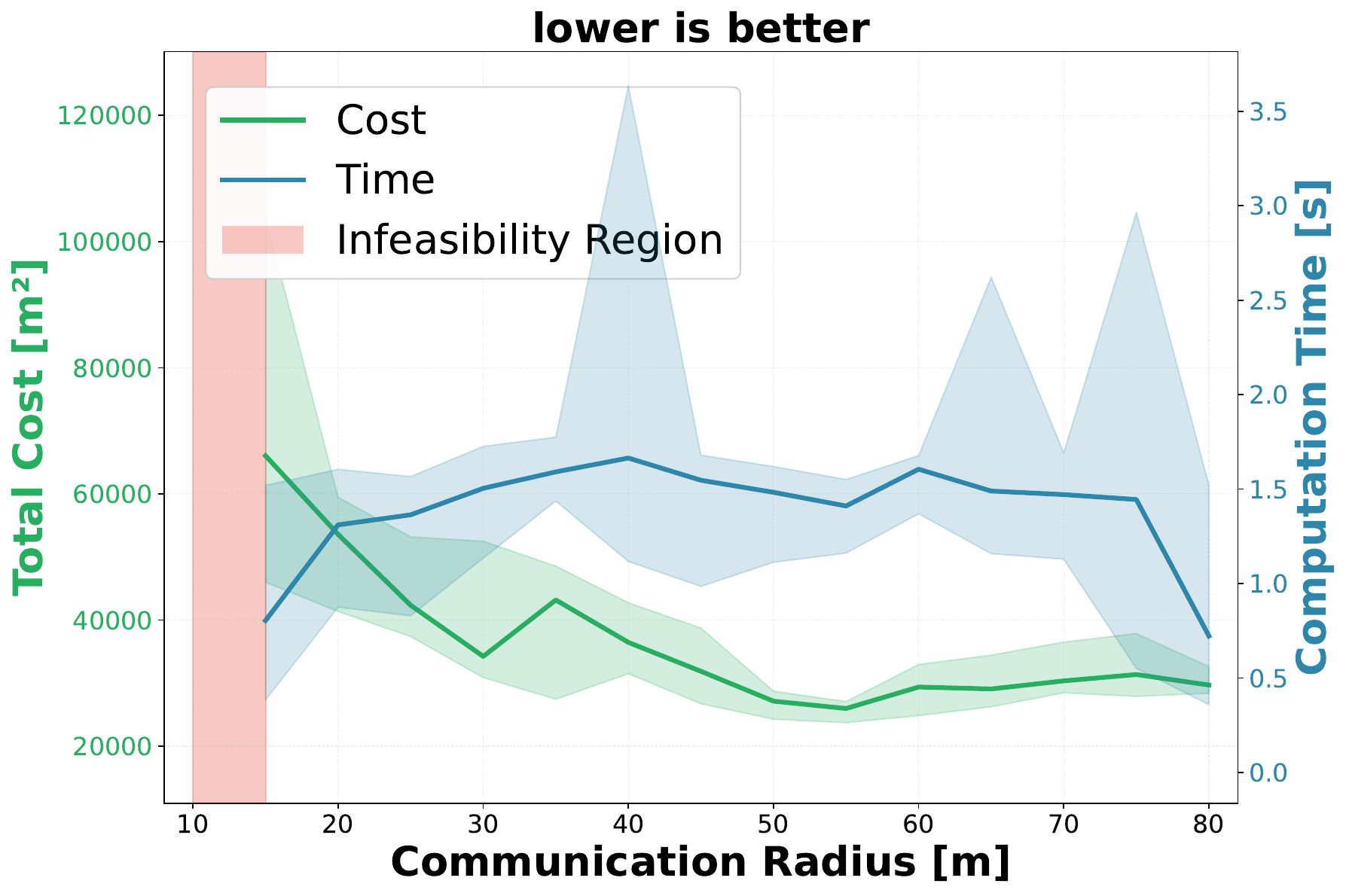}
         (a) Varying $r_{\text{max}}$ with $r_{\text{comm}} =$ \qty{55}{\meter} fixed. (b) Varying $r_{\text{comm}}$ with $r_{\text{max}} =$ \qty{40}{\meter} fixed.
         \label{fig:sensitivity_rcomm}
    \caption{\small Sensitivity analysis showing coverage cost and runtime varying (a) maximum disk radius $r_{\text{max}}$ and (b) communication radius $r_{\text{comm}}$. Lines show median over 10 trials on 10 problem instances; shaded regions show min-max range. Red shaded regions indicate parameter ranges where the algorithm fails to achieve coverage requirements ($n=200$ assets, $m=50$ robots in a \qty{100}{\meter} $\times$ \qty{100}{\meter} workspace, coverage requirements $\kappa(p)$ uniformly sampled from $\{1,2,3\}$).}
    \label{fig:sensitivity}
\end{figure}

We analyze the algorithm's sensitivity to communication radius $r_{\text{comm}}$ and maximum sensing radius $r_{\text{max}}$ in Fig.~\ref{fig:sensitivity}. When varying $r_{\text{comm}}$ with fixed $r_{\text{max}} =$ \qty{40}{\meter} (Fig.~\ref{fig:sensitivity}(b)), the algorithm fails below $r_{\text{comm}} =$ \qty{15}{\meter} due to insufficient coordination between robots. As $r_{\text{comm}}$ increases, total cost decreases as robots gain better knowledge of neighboring coverage, reducing unnecessary disk growth. Conversely, when varying $r_{\text{max}}$ with fixed $r_{\text{comm}} =$ \qty{55}{\meter} (Fig.~\ref{fig:sensitivity}(a)), the algorithm fails below $r_{\text{max}} =$ \qty{20}{\meter} due to physical infeasibility. Beyond this threshold, cost increases with $r_{\text{max}}$ due to a communication-sensing mismatch: robots can sense and cover distant assets up to $r_{\text{max}}$ away, but can only communicate within $r_{\text{comm}}$, causing them to perceive already-covered distant assets as undercovered and grow unnecessarily large disks.
While $r_{\text{comm}}$ and $r_{\text{max}}$ are determined by sensor specifications, algorithmic parameters $\epsilon = 0.01$ (tie tolerance), $\tau = 0.005$ (swap threshold), chosen empirically, prove robust across instances. For new deployments: use hardware values for $r_{\text{comm}}$, $r_{\text{max}}$, and tune $\epsilon, \tau$.

\section{Discussion and Conclusion}\label{sec:discussion}

We presented a distributed multi-coverage algorithm for robot swarms operating under local sensing ($r_{\max}$), local communication ($r_{\text{comm}}$), and no global coordination. Through systematic exploration, marginal cost optimization, and iterative refinement, the algorithm satisfies heterogeneous coverage requirements $\kappa(p)$ while maintaining near-constant computation time across problem scales. 

Key limitations include: (1) sensitivity to parameter selection—the algorithm fails when $r_{\text{comm}} <$ \qty{15}{\meter} or $r_{\text{max}} <$ \qty{20}{\meter} for our test instances, and requires $r_{\text{comm}} \geq r_{\text{max}}$ to avoid coordination failures; (2) increasing optimality gap with larger swarm sizes due to coordination complexity; and (3) potential for persistent overcoverage in Phase 3 refinement, as local neighbor information cannot eliminate all unnecessary redundancy achievable through global optimization.

Despite these limitations, the 45.78\%--69.41\% optimality gap is acceptable for applications prioritizing responsiveness (12--410× speedup), operation under communication constraints ($r_{\text{comm}}$ < workspace), or dynamic adaptation (29× faster in Fig.~\ref{fig:dynamic_comparison}). These tradeoffs make our algorithm suitable for deployments where local constraints preclude centralized control and responsiveness outweighs solution optimality. Centralized methods remain preferable for offline planning with fixed assets, cost-constrained deployments justifying longer computation, or small problems solvable in seconds.

Future work includes theoretical analysis of optimality bounds and hardware experiments with failing robots. Handling infeasible parameter regions could involve graceful degradation strategies that prioritize critical assets while relaxing $\kappa(p)$ for lower-priority assets. Extending the algorithm to asynchronous execution, where robots update at different rates, would require timestamped state exchange and conservative bid validation to maintain correctness, likely increasing iteration counts but enabling deployment on systems without global clock synchronization. Our marginal cost bidding framework generalizes to distributed task allocation and formation control, providing a template for distributed algorithms via conservative coordination.

\begin{credits}
\subsubsection{\ackname}  This work was supported by the Army Research Laboratory under Cooperative Agreement Number W911NF-23-2-0014. The views and conclusions contained in this document are those of the authors and should not be interpreted as representing the official policies, either expressed or implied, of the Army Research Laboratory or the U.S. Government. The U.S. Government is authorized to reproduce and distribute reprints for Government purposes notwithstanding any copyright notation herein.

\subsubsection{\discintname}
The authors have no competing interests to declare that are
relevant to the content of this article.
\end{credits}
%
%
%
\clearpage  
\bibliographystyle{splncs04}
 
\bibliography{mybib}

\end{document}